# AN EFFECTIVE MIXTURE-OF-EXPERTS APPROACH FOR CODE-SWITCHING SPEECH RECOGNITION LEVERAGING ENCODER DISENTANGLEMENT

*Tzu-Ting Yang*★*, Hsin-Wei Wang*★*, Yi-Cheng Wang*★*, Chi-Han Lin*†*, and Berlin Chen*★

★National Taiwan Normal University, Taipei, Taiwan
†E.SUN Financial Holding Co., Ltd.
{tzutingyang, hsinweiwang, yichengwang, berlin}@ntnu.edu.tw

**Abstract**

With the massive developments of end-to-end (E2E) neural networks, recent years have witnessed unprecedented breakthroughs in automatic speech recognition (ASR). However, the code-switching phenomenon remains a major obstacle that hinders ASR from perfection, as the lack of labeled data and the variations between languages often lead to degradation of ASR performance. In this paper, we focus exclusively on improving the acoustic encoder of E2E ASR to tackle the challenge caused by the code-switching phenomenon. Our main contributions are threefold: First, we introduce a novel disentanglement loss to enable the lower-layer of the encoder to capture inter-lingual acoustic information while mitigating linguistic confusion at the higher-layer of the encoder. Second, through comprehensive experiments, we verify that our proposed method outperforms the prior-art methods using pre-trained dual-encoders, meanwhile having access only to the code-switching corpus and consuming half of the parameterization. Third, the apparent differentiation of the encoders' output features also corroborates the complementarity between the disentanglement loss and the mixture-of-experts (MoE) architecture.

## 1. INTRODUCTION

Building upon the widespread adoption of end-to-end (E2E) neural networks, automatic speech recognition (ASR) models have made great strides and achieved conspicuous success in many general use cases. Contrary to classical ASR that is based on the decomposition into acoustic model, language model and more, E2E ASR integrates all available information within a unified neural network model, demonstrating promising performance on many benchmark tasks. On a separate front, due to the rapid proliferation of social media, ASR has become the best preferred option for many creators to generate transcriptions or closed captions for multimedia content access. Today, more than 60% of the world's population are fluently multilingual speakers [1], while most of them often unconsciously use different languages interleavingly in daily conversations. This leads to the so-called code-switching (CS) phenomenon. Despite that many research efforts have pushed the limit of E2E ASR [2][3][4][5], making it reach human parity for monolingual use cases, whose performance often degrades dramatically when dictating code-switching conversations. This therefore creates a pressing need for research on code-switching ASR [6][7][8].

The major difficulty incurred when handing the code-switching phenomenon is the lack of sufficient labeled data accompanied with vast divergence of phonological and syntactic structures among different languages [9]. When these different language traits are intertwined, E2E ASR models often struggle to learn the acoustic and lexical knowledge of a specific language pertinently in a principal manner. It is more exacerbated for the language pair of Mandarin and English. In contrast to most European languages like English, the Mandarin language consists of many pictographic characters rather than alphabet letters, where a sizable number of homophones and heterogeneous characters, words and phrases frequently occur. In addition, tonal information plays a crucial role in the choice of characters for Mandarin, apart from emphasizing the specific emotions of given words or expressions [10].

In order to reduce the processing complexity of code-switching ASR, it is intuitive but crucial to integrate the information about language identification (LID), which makes an ASR model aware of the language being transcribed [11][12][13][14]. Such an integration facilitates subsequent modules to focus more on the transcription of the current specific language while reducing inter-lingual contextual confusion. For example, previous studies [10][15][16][17] manage to add a decoder for predicting language types to handle the issue of language confusion, guiding the ASR model to build awareness of language differences through multi-task training [18]. Yet another study [19] learns the domain-specific knowledge pertaining to each language in isolation with the self-attention mechanism, which masks the text embedding at the decoder with a prior classification objective according to LID. On the other direction, [1][20] suggest that incorporating individual encoders for each language can simplify the effort on extracting domain knowledge. However, other researchers [21] caveat that having dual-encoders may make an ASR model overloaded with too many parameters. Furthermore, independently training the encoders may cause inter-lingual information cues such as synonyms and interleaving patterns flushed away from the model. Because of these, the language-aware encoder (LAE) with a shared encoder block was proposed [21][22].

In response, we aim to increase ASR robustness in scenarios with a frequent code-switching phenomenon through improved modeling of the acoustic encoder. Our contributions can be summarized as follows: 1) we introduce a disentanglement loss to enables the lower-layer of the encoder to render inter-lingual acoustic information while mitigating linguistic confusion at the higher-layer of the encoder; 2) through experiments, we verify that our proposed method outperforms the conventional methods, meanwhile having access only to the code-switching corpus and consuming half of the parameterization; and 3) based on the apparent differentiation of the encoders' output features, we confirm the complementarity between the disentanglement loss and the mixture-of-experts (MoE) architecture.

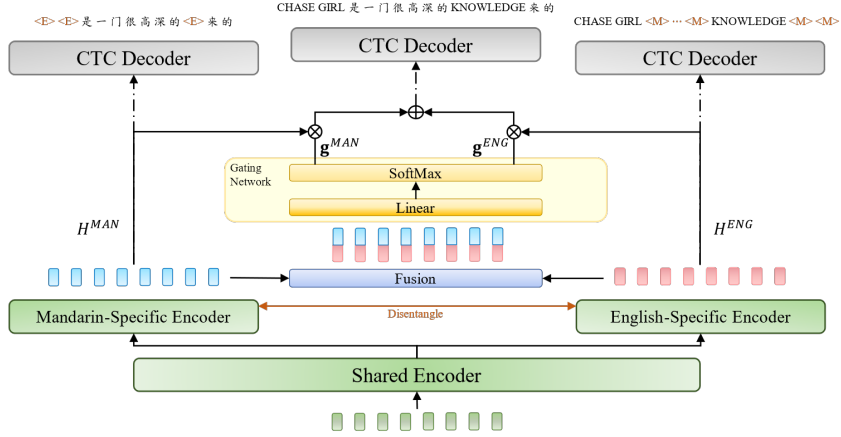

*Figure 1: Model architecture of the proposed mixture-of-experts based disentanglement language-aware model.*

The rest of this paper is organized as follows. Section 2 sheds light on our methodology, followed by Section 3 that provides in-depth analysis on the corresponding experiments, respectively. Finally, Section 4 concludes the paper and suggests future research avenues.

## 2. METHODOLOGY

### 2.1. Language-Aware Encoder (LAE)

The overall architecture of our proposed mixture-of-experts based disentanglement language-aware model is illustrated in Figure 1. We choose the LAE as our as our archetype, which uses the lower encoder layers as a shared encoder block, compared to the traditional dual-encoder architecture [21]. The shared encoder decreases a significant number of model parameters and retains the inter-lingual acoustic information by initially co-processing the acoustic representation. At the outset, the acoustic representation input $\mathbf{X} = (\mathbf{x}_l \in \mathbb{R}^D | l = 1, \cdots, L)$ is fed into the shared encoder to extract the inter-lingual acoustic feature embedding $\widetilde{\mathbf{H}}$:

$$\widetilde{\mathbf{H}} = Shared\_Encoder(\mathbf{X}). \tag{1}$$

Subsequently, the language-specific encoders will proceed to extract high-level language-dependent semantic information:

$$\mathbf{H}^{MAN} = Specific\_Encoder^{MAN}(\widetilde{\mathbf{H}}), \tag{2}$$

$$\mathbf{H}^{ENG} = Specific\_Encoder^{ENG}(\widetilde{\mathbf{H}}). \tag{3}$$

Previous studies [21][22] have employed language-aware target (LAT) for learning language-specific encoders, enabling the model to discriminate different language sources with respect to reference (oracle) output $\mathbf{y} = (y_s \in V | s = 1, \cdots, S)$. Mandarin LAT $\mathbf{y}^{MAN}$ is the reference output with its English part being masked, and vice versa for English LAT $\mathbf{y}^{ENG}$. For example, $\mathbf{y}^{MAN}$ is set to contain only Mandarin characters intertwined with language masks <English> as the target output for training the Mandarin-specific encoder. We utilize both $\mathbf{y}^{MAN}$ and $\mathbf{y}^{ENG}$ to compute the auxiliary language-specific loss $\mathcal{L}_{Lang}$ via CTC [23], encouraging the language-specific encoders to focus exclusively on monolingual learning:

$$\mathcal{L}_{MAN} = CTC(\mathbf{y}^{MAN}, \mathbf{H}^{MAN}), \tag{4}$$

$$\mathcal{L}_{ENG} = CTC(\mathbf{y}^{ENG}, \mathbf{H}^{ENG}), \tag{5}$$

$$\mathcal{L}_{Lang} = \frac{\mathcal{L}_{MAN} + \mathcal{L}_{ENG}}{2}. \tag{6}$$

### 2.2. Integration into the MoE Architecture

We notice that merely adding [21] or concatenating [22] the output features of these expert models may not fully utilize their capabilities. [24] introduce the notion of MoE to enhance the dual-encoder architecture that simply consists of two independent pre-trained encoders, in contrast to LAE. Specifically, MoE comprises two major components: two expert models and a gating network. The expert models focus on learning language-specific embeddings individually. The gating network evaluates the importance of each expert and dynamically generates the frame-level language weights. To bring the MoE architecture into the LAE model, we first concatenate the $\mathbf{H}^{MAN}$ and $\mathbf{H}^{ENG}$ into $\mathbf{H}^{concat}$ and use it as input to the gating network:

$$\mathbf{G} = Gating\_Network(\mathbf{H}^{concat}). \tag{7}$$

Then the output $\mathbf{G}$ is fed into $Softmax(\cdot)$:

$$P(\mathbf{g}^{MAN}, \mathbf{g}^{ENG} | \mathbf{H}^{concat}) = Softmax(\mathbf{G}), \tag{8}$$

where both $\mathbf{g}^{MAN}$, $\mathbf{g}^{ENG}$ are $\mathbb{R}^L$ frame-level language weights. Subsequently, the frame-level output $\mathbf{H}^{MoE}$ of MoE can be calculated by:

$$\mathbf{H}^{MoE} = \mathbf{g}^{MAN} \cdot \mathbf{H}^{MAN} + \mathbf{g}^{ENG} \cdot \mathbf{H}^{ENG}. \tag{9}$$

After using $\mathbf{H}^{MoE}$ to calculate the mixture loss $\mathcal{L}_{Mix}$. The total training objective is defined as follows:

$$\mathcal{L}_{Mix} = CTC(\mathbf{y}, \mathbf{H}^{MoE}), \tag{10}$$

$$\mathcal{L}_{Total} = \frac{1}{2}(\mathcal{L}_{Mix} + \mathcal{L}_{Lang}) + \lambda \, \mathcal{L}_{Disentangle}, \tag{11}$$

where $\lambda$ is coefficients that is appropriately set to keep the antecedent and consequent terms in Eq. (11) at a similar scale. Notably, $\mathcal{L}_{Lang}$ is only utilized in the training period and does not bring additional computation in the inference period.

*Table 1: The SEAME corpus was analyzed in terms of the number of speakers, the total duration, and the Mandarin (MAN)-to-English (ENG) ratio.*

|  | Train | Valid | Dev$_{MAN}$ | Dev$_{SGE}$ |
|---|---|---|---|---|
| Speakers | 134 | 134 | 10 | 10 |
| Duration (hrs) | 97.9 | 5.2 | 7.5 | 3.9 |
| MAN/ENG | 68/32 | 67/33 | 74/26 | 37/63 |

### 2.3. Disentangling Between Two Languages

Most notably, we expect the two different encoders as the expert models and capture disparate language-specific semantic information. While the lower-layer shared encoder block can extract inter-lingual acoustic information for the expert models, it may also lead to the linguistic confusion between the two expert models. Therefore, we expect that incorporating the disentanglement loss in the output of the two expert models can effectively address high-level linguistic confusion meanwhile leveraging lower-level inter-lingual information. We use the cosine-distance ($CD$) between $\mathbf{H}^{MAN}$ and $\mathbf{H}^{ENG}$ as a basis to form the disentanglement loss:

$$CD(\mathbf{h}_{i,j}^{MAN}, \mathbf{h}_{i,j}^{ENG}) = 1 - \frac{\mathbf{h}_{i,j}^{MAN} \cdot \mathbf{h}_{i,j}^{ENG}}{\|\mathbf{h}_{i,j}^{MAN}\|_2 \|\mathbf{h}_{i,j}^{ENG}\|_2}, \quad (12)$$

$$\mathcal{L}_{CD} = -\frac{1}{N}\sum_{i=1}^{N}\frac{1}{|s_i|}\sum_{j=1}^{|s_i|} CD(\mathbf{h}_{i,j}^{MAN}, \mathbf{h}_{i,j}^{ENG}), \quad (13)$$

where $N$ is the number of training utterances and $s_i$ is the $i^{th}$ training utterance; $\mathbf{h}_{i,j}^{MAN}$ is the $j^{th}$ hidden embedding of $s_i$ in $\mathbf{H}^{MAN}$, and vice versa. When the cosine-distance increases, the disentanglement between $\mathbf{h}_{i,j}^{MAN}$ and $\mathbf{h}_{i,j}^{ENG}$ is more pronounced.

## 3. EXPERIMENTS

### 3.1. Experimental Setup

All experiments are conducted on the SEAME [25] dataset, a spontaneous code-switching corpus recorded by Southeast Asian speakers. Both intra-sentence and inter-sentence code-switching speech utterances exist in SEAME. The total duration of recorded audio is about 115 hours, and the corresponding detailed statistics is shown in Table 1.

We adopted Transformer CTC as the backbone architecture. More specifically, in the baseline model, we used 15 Transformer blocks to constitute the encoder, where the feedforward dimension, attention dimension, and attention head were set to 2048, 256 and 4, respectively. Sine this paper aims primarily to enhance the encoder of the ASR model, so we adopt the commonly-used CTC as the decoder. To ensure a fair comparison between our modeling method and the baseline system, we set the number of Transformer blocks contained in the shared encoder block, the Mandarin-specific encoder and the English-specific encoder in LAE to 9, 3 and 3, respectively. For language model, we used 4 Transformer blocks with the same dimension as the encoder. The gating network contains only a single linear layer, which minimizes its impact on the number of model parameters. All the reference text transcripts of the training utterances are converted into the corresponding 2,624 Mandarin characters, 3,000 English BPEs and two language masks, namely, <Mandarin> and <English>. We set the hyperparameter λ in Eq. (11) to 10. The Adam optimizer was used to perform 100 training epochs with an initial warm-up phase of 25,000 steps. Then the final model was obtained by averaging the top 10 checkpoints based on their validation scores. When beam search is enabled, we set the beamwidth to 10. For the final evaluation, we use the mixed error rate (MER) that comprises the character error rate (CER) for Mandarin and the word error rate (WER) for English.

### 3.2. Overall Comparison

While both Bi-Encoder and LAE have shown remarkable performance on the ASRU2019 corpus (recorded by East Asian speakers) [28], they seem to have not performed as well as expected at SEAME. The ratio of English to Mandarin is only about 10% for ASRU2019. However, as depicted in Table 1, the statistics indicate that at least 30% of the reference transcripts in the SEAME corpus consist of English words. It is reasonable to assert SEAME contains more diverse code-switching phenomena than ASRU2019.

As such, we compare the performance of our proposed method to several recently novel and relevant modeling structures on SEAME, as shown in Table 2. With prior refinement of the encoder by pre-training on Aishell-1 [26] and Librispeech_clean_100 [27], Bi-Encoder shows marginal improvements on both the Dev$_{MAN}$ and Dev$_{SGE}$ test sets. Moreover, LAE captures inter-lingual information through a shared encoder block, resulting in a MER reduction of 0.2% on Dev$_{SGE}$ with only half of the parameters. Our proposed method, the MoE-based language-aware disentanglement model, represents an enhancement of LAE can further improve the overall performance. As shown in Table 1, for Dev$_{MAN}$ and Dev$_{SGE}$, the ratio of Mandarin to English is diametrically opposite to each other. Evidently, Dev$_{SGE}$ contains more frequent code-switching phenomenon, which poses a more challenging task for the model. Our proposed method stands out for Dev$_{SGE}$, confirming that the disentanglement loss can effectively mitigate potential semantic confusion caused by the shared encoder block at the lower-layer of the encoder.

To facilitate a comparison between our method with another recent method, Multi-Transformer-Transducer [1], we further incorporate a shallow language model fusion with LSTM into our model structure. Despite better performance is obtained in relation to Multi-Transformer-Transducer, the improvement on Dev$_{SGE}$ is less pronounced than expected. This may be attributed that unlike the built-in language model adopted by the former, only a simple shallow fusion is employed in our model structure language model.

*Table 2: Compare the MER results of our method with Transformer CTC, Bi-Encoder, LAE and Multi-Transformer-Transducer.*

| Model | LM | MER (%) | | Params. |
|---|---|---|---|---|
| Transformer CTC | ✗ | Dev$_{MAN}$ | 21.4 | 23.05 M |
| | | Dev$_{SGE}$ | 30.3 | |
| Bi-Encoder [24] | ✗ | Dev$_{MAN}$ | 21.0 | 44.58 M |
| | | Dev$_{SGE}$ | 29.7 | |
| LAE [21] | ✗ | Dev$_{MAN}$ | 21.0 | 24.46 M |
| | | Dev$_{SGE}$ | 29.5 | |
| Proposed Method | ✗ | Dev$_{MAN}$ | **20.7** | 24.46M |
| | | Dev$_{SGE}$ | **29.0** | |
| Multi-Transformer-Transducer [1] | ✓ | Dev$_{MAN}$ | 20.2 | - |
| | | Dev$_{SGE}$ | 27.7 | |
| Proposed Method | ✓ | Dev$_{MAN}$ | **19.2** | - |
| | | Dev$_{SGE}$ | **27.1** | |

*Table 3: The performance of our proposed method is evaluated in terms of Mandarin (CER), English (WER), and MER on the test sets Dev$_{MAN}$ and Dev$_{SGE}$ under different settings.*

|  |  | Concatenated-LAE | | | MoE based LAE | | |
|---|---|---|---|---|---|---|---|
|  |  | Mandarin CER (%) | English WER (%) | CS MER (%) | Mandarin CER (%) | English WER (%) | CS MER (%) |
| w/o Disentangle | Dev$_{MAN}$ | 18.5 | 38.2 | 21.1 | 20.6 | 42.6 | 23.6 |
|  | Dev$_{SGE}$ | 27.5 | 36.1 | 29.8 | 30.0 | 39.5 | 32.4 |
| w Disentangle | Dev$_{MAN}$ | 18.5 | 38.2 | 21.1 | **18.3** | **37.3** | **20.7** |
|  | Dev$_{SGE}$ | 27.5 | 35.9 | 29.7 | **26.7** | **35.3** | **29.0** |

### 3.3. Component-wise Evaluation

In the following experiments, we will implement a Concatenated-LAE modification, by concatenating the embedding obtained from Mandarin and English experts as an alternative fusion mechanism. As shown in Table 3, the performance gain brought by the disentanglement loss seems to be inconspicuous when with Concatenated-LAE. This is because the outputs of both experts being concatenated are essentially in two different latent representation spaces, so it is acceptable that there is no improvement in disentangling these two features.

Unexpectedly, when MoE works alone without imposing the disentanglement loss, the MER results are surprisingly increased by more than 10% compared to Concatenated-LAE. Due to the expert models generating similar frame-level features, the incorporation of the gating network becomes disoriented when evaluating the importance of each expert model. In this case, mixture-of-experts architecture not only increases the complexity of the model during the training process but also makes the model increasingly confused about the correlations between the two languages.

Through our experiments, it is evident that the disentanglement LAE and the MoE architecture are complementary to each other. The disentanglement LAE allows LAE to reducing semantic confusion at the higher-layer of the encoder while effectively leveraging the inter-lingual acoustic information by the shared encoder. MoE architecture treats both language-specific encoders as expert models and dynamically allocates their weights to leverage their potential fully. More specifically, after introducing the disentanglement loss, our proposed method, MoE based disentanglement LAE, achieves an absolute MER reduction of 2.9% on Dev$_{MAN}$ and 3.4% on and Dev$_{SGE}$.

### 3.4. Analyze the Latent Representation

To verify the effectiveness of our proposed method, we depict the output embeddings of the experts utilizing t-SNE method [29]. Figures 2(a) and 2(b) show the 2D embedding space before and after disentangling, respectively. We can find some overlap of output embeddings between different experts before disentangling, which implies confusion on the semantic level. Our method disentangles the language-specific features by maximizing the cosine distance between them to reduce their similarity. This process is effective in reducing semantic confusion in the higher-layer of the encoder.

In order to know whether the outputs of the experts are over-differentiated, we further analyze their mixture weights. Figures 2(c) and Fig. 2(d) show the output weights of the gating network before and after disentangling, respectively. Considering the dominance of Mandarin in SEAME, the gated network prefers to increase the weight of the Mandarin expert feature. Through the application of the disentanglement loss, when the model encounters English words, the weight of English experts is effectively raised. This indicates that the model is more confident in discriminating the language. It is worth mentioning that SEAME contains a large number of discourse particles, and the discourse particles with same pronunciation may have both Mandarin and English token labels [30]. Therefore, the model that relies only on the predictions of the acoustic encoder may be error-prone when determining the language type of the auxiliary "AH" at the end of the example, but such errors are inevitable.

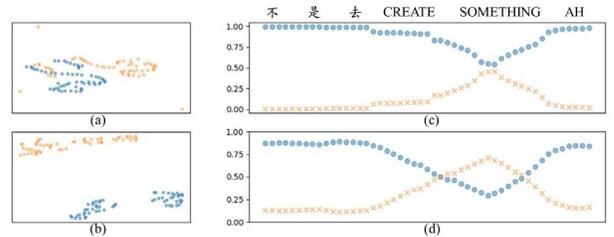

*Figure 2: We project the outputs of the Mandarin expert (blue) and the English expert (orange) to the 2D plane using the t-SNE algorithm and visualize their outputs weights generated by the gating network. (a) and (b) denote the embedding projection maps in 2D space obtained by using t-SNE; (c) and (d) are the diagrams containing language-specific weights.*

### 4. CONCLUSIONS

In this paper, we have proposed an effective modeling method for Mandarin-English code-switching ASR, which leverages language-aware encoder trained with novel disentanglement loss and gating mechanism. A series of experiments carried out on the SEAME benchmark dataset have demonstrated the effectiveness of our method in relation to some state-of-the-art baselines, confirming the utility and practicality of our method for code-switching ASR. As for future work, we plan to pair our method with more sophisticated neural structures and investigate in-depth a principal way to model the intricate Mandarin-English code-switching phenomena for robust ASR.

### 5. ACKNOWLEDGEMENT

This work was supported in part by E.SUN Bank under Grant Number 202308-NTU-02. Any findings and implications in the paper do not necessarily reflect those of the sponsors.